# The power of context: Random Forest classification of near synonyms. A case study in Modern Hindi

*Jacek Bąkowski*

*Institute of Polish Language, Polish Academy of Sciences, Poland*

**Abstract**

Synonymy is a widespread yet puzzling linguistic phenomenon. Absolute synonyms theoretically should not exist, as they do not expand language's expressive potential. However, it was suggested that even if synonyms denote the same concept, they may reflect different perspectives or carry distinct cultural associations, claims that have rarely been tested quantitatively.

In Hindi, prolonged contact with Persian produced many Perso-Arabic loanwords coexisting with their Sanskrit counterpart, forming numerous synonym pairs. This study investigates whether centuries after these borrowings appeared in the Subcontinent their origin can still be distinguished using distributional data alone and regardless of their semantic content.

A Random Forest trained on word embeddings of Hindi synonyms successfully classified words by Sanskrit or Perso-Arabic origin, even when they were semantically unrelated, suggesting that usage patterns preserve traces of etymology. These findings provide quantitative evidence that context encodes etymological signals and that synonymy may reflect subtle but systematic distinctions linked to origin. They support the idea that synonymous words can offer different perspectives and that etymologically related words may form distinct conceptual subspaces, creating a new type of semantic frame shaped by historical origin. Overall, the results highlight the power of context in capturing nuanced distinctions beyond traditional semantic similarity.

# 1. Linguistic Background

## 1.1. The problem of synonymy

Synonymy is a commonly occurring and widespread, but also intriguing, linguistic phenomenon. Apart from stylistic considerations, such as avoiding repetition, there is no semiotic justification for the existence of absolute synonyms. Indeed, the economy of language suggests that no language permits a perfect fit between two words in all respects. Maintaining a double vocabulary with many synonyms violates Zipf's principle, according to which the tendency to implement speech phenomena with the least effort plays a major role in the evolution of language.

Moreover, "instead of increasing the expressive potential of the language by making a single lexical item express multiple meanings, near-synonymy decreases the language's expressive power by allowing several items to convey the same meaning" (Divjak 2010: 1). Hence, it is claimed that "total synonymy is an extremely rare occurrence, a luxury that language can ill afford" (Ullmann 1957: 108) and it seems to be the consensus in the field of semantics that natural languages „abhor absolute synonyms just as nature abhors a vacuum" (Cruse 1986: 270).

Apart from sphere of slang and informal speech, where a significant proliferation of synonyms can be observed, in formal language synonymy is often limited to pairs of foreign and native terms. In practice, even if a loanword is adopted with exactly the same meaning as an existing word (thus creating a true synonym), the two words quickly diverge—not necessarily in the definition itself, but they differ in usage, shade of meaning, register, social meaning or all of these: thus developing a difference in their pragmatic function. As a consequence, despite semantic proximity, they exhibit different connotational meaning, stylistic distribution, and frequency of occurrence and thus, occupy a slightly different semantic area (Urdang, Laroche 1978: iii). Moreover, synonyms may also carry different cultural stereotypes.

Thus, from a functional perspective, the existence of synonyms is considered an aberration, and if absolute synonyms exist at all, they are extremely uncommon. This leads to the *principle of no synonymy* (Goldberg 1995: 67), recently refined as the *principle of no equivalence* (Leclercq & Morin 2023), which holds that any difference in form should entail a difference in meaning.

In such a cognitive approach a word is a world and its meaning enriched with an element of psychology taking into account the concepts and values associated with it in a given culture and language. It has been suggested that "even if near-synonyms do name one and the same thing, they name it in different ways: they present different perspectives on a situation" (Divjak 2010: 1). Although this claim sounds convincing, it has never been thoroughly investigated on a large scale using quantifiable methods.

One of the purposes of this paper is to partially fill that gap and explore the differences in perception implied by synonymy using computational linguistic methods.

## 1.2. The origin of loanwords in India

Synonyms often come from different historical layers that make up a single language. Loanwords can be a rich source of synonyms, often from languages perceived as more prestigious by association with the dominant culture in a region. This is the case in most European languages which have borrowed from Latin or ancient Greek, but where the native terms continue to be used in parallel.

Such is also the case in India: it has never been isolated, and from the very beginning of its existence has been subject to foreign influence. "Jutting into the heart of the Indian Ocean—the world's oldest maritime zone—and connected to the Iranian plateau by several strategic mountain passes, the Indo-Gangetic plain and the great peninsula to its south have long been a major crossroads of transregional movement and exchange. Pathways leading to and across the subcontinent have carried a wide range of global flows, while migrating populations brought or took away diverse cultural traditions embracing statecraft, architecture, warfare, cuisine, religion and what is interesting to us—language" (Eaton 2019: 380).

Although this process has very deep roots, it accelerated between the 11th and 18th centuries, when two transregional worlds came into close contact: Sanskrit and Persian. During this time Persianate culture spread across almost all of the Indian subcontinent and the Persian language interacted with its Sanskrit counterpart. The history of this interaction is both rich and complex. Suffice it to say that by the time Persian had become a vibrant and prestigious literary language, its knowledge became a practical necessity, whether one was seeking a literary career, or a position in the state administration. But knowledge of Persian was also a matter of prestige: as a language of the ruling class, it became the language of chivalry, war and love: it served as a guide to accessing court culture, and played the role of a transregional medium of culture and education. For a few centuries India became a major centre of the Persian speaking or reading world (Eaton 2019: 8, 380–2; Saksena 1927: 3–5; Kuczkiewicz-Fraś 2012: 47-51).

## 1.3. The peculiarity of Modern Hindi

Such long and intense linguistic contact, spanning a period of about seven centuries, must clearly have left a mark on India's linguistic landscape.

Although, from the 18th century, the importance of Persian in India began to slowly decline with the gradual weakening of the Mughal empire (in its place came English), the legacy of the language contact with Persian remains vivid in the languages of South Asia — particularly in northern and western vernacular languages of the subcontinent, including Modern Hindi.

The initially strange and foreign Perso-Arabic vocabulary was assimilated and integrated into the fabric of Hindi. However, new elements borrowed from Persian did not replace native ones but started "functioning side by side, becoming in the course of time a rightful ingredient of the whole system of Hindi, enriching significantly the Hindi lexis" (Kuczkiewicz-Fraś 2003: 10). Today, Persian „still provides the Indo-Aryan languages with their most substantial set of non-Indian elements" (Kuczkiewicz-Fraś 2012: 56) and words of Perso-Arabic and Sanskrit origin are often used concurrently as synonyms.

This coexistence of Sanskrit-derived and Perso-Arabic elements within Hindi can be described as a lexical continuum where the two layers of vocabulary interpenetrate and complement one another, diverging primarily in stylistic, cultural, and religious associations. The choice between them depends on context, register, and the current communicative situation. Sometimes, a word borrowed from Perso-Arabic is preferred over its unnatural-sounding Sanskrit synonym, whereas in other contexts, it would sound odd. However, it seems that these patterns of lexical variation have never been systematically examined from the perspective of computational linguistics.

## 2. Context and Distributive Semantics

### 2.1. The Context and the Meaning

Meaning is an elusive phenomenon. It is "far more complex than grammar, and far more difficult to study and describe", but also "complete semantic descriptions cannot realistically be envisaged". Consequently, any semantic description "must limit itself to facets of the total meaning" (Langacker 2018: 11).

Even a single encounter with a word can potentially leave a memory trace of its use. Word meanings evoke rich conceptual and perceptual information derived from the contexts in which they have been previously witnessed. Beneath conscious awareness, we record not only the forms of utterances, but also the concepts and interpretations associated with them, along with the contexts in which they were heard or seen — each adding a subtle shade of meaning. Elements of context that are perceived as more relevant or specific to a given word are more likely to be encoded in a speaker's implicit memory. Contexts associated with a word are continuously updated, and additional encounters can overlap over time, making some aspects more central to a word's meaning (Goldberg 2019: 11–27).

Furthermore, "what people know when they know a word is not how to recite its dictionary definition—they know how to use it in everyday discourse [...]. Knowing how to use words is a basic component of knowing a language" (Miller & Charles 1991: 4). Thus, what we know when we know a word is its context. Context has become a key element not only in how we conceive language and its speakers' knowledge, but also a major factor in semantic change. Indeed, language is a dynamic system emerging from use, where "language change is change in use" (Traugott 2010: 30) and semantic shifts are usually originated by a change in the typical context of a word (Tang 2018).

Context is thus a key concept for understanding the meaning of words and it appears to be one of the fundamental ingredients of human conceptualization (Lenci 2008: 5). Therefore, it should play a crucial role in any realistic account of language cognition (Divjak 2019: 6).

## 2.2. The importance of distributive semantics

For years, a context and usage-based approach to meaning has been gaining increasing attention in both linguistics and cognitive science. This approach to semantics brings computational linguistics to the fore, and particularly word embedding techniques, which allow us to extract information about word similarities from large language corpora.

Since a corpus is a large collection of produced texts or spoken utterances that provides a representative sample of actual language use, it can be assumed to represent—to some extent—the language. One implication of this approach is the hypothesis that word co-occurrence statistics, extracted from large text corpora, form a basis for semantic representation, and that the statistical distribution of words in context can be a key element in understanding some of their semantic characteristics.

In other words, "by inspecting a significant number of linguistic contexts representative of the distributional and combinatorial behavior of a given word, we may find evidence about (some of) its semantic properties" (Lenci 2008: 3). It does not matter here whether we conceive this functional dependence to be merely correlational or a truly causal relation; however, the present author's conviction is that the two are connected: words occur in a certain context because they have a certain meaning, but that meaning is secondarily enriched by common usage patterns.

## 2.3. The distributional hypothesis

The degree of semantic similarity between two lexemes can be modeled as a function of the degree of overlap between their linguistic contexts (Harris 1954; Miller & Charles 1991; Baroni & Lenci 2010). In other words, lexemes that share many linguistic contexts are probably semantically similar, which brings the discussion back to the classic formulation: "You shall know a word by the company it keeps" (Firth 1957). Moreover, the particular contexts in which different words appear can reveal—again, to some extent—the semantic drift between terms that are otherwise considered synonyms.

Thus, the question arises whether, several centuries after Perso-Arabic borrowings appeared in Hindi, is it possible to perceive differences between closely related near synonyms using distributional semantics, that is basing on their context and regardless of their near-identical meaning. And whether it is attainable, with machine learning tools, to detect usage patterns which would allow to classify them.

# 3. Methodology

## 3.1. Selection of synonymous pairs

For the purpose of this paper, 135 pairs of synonyms of Sanskrit/Perso-Arabic origin were selected, which provides us 270 words (see the code repository for a full list). There is no objective methodology for selecting such pairs, nor is there a neat way of characterizing synonyms (Cruse 1986: 266), since the subtle nuances of meaning among them remain unclear (Geeraerts 2010: 85). Thus, one must partially rely on our intuition. The task is complicated by the fact that although there are many synonymous pairs in Hindi, dictionaries sometimes show large discrepancies in their definitions.

An additional challenge arises from the need to avoid homonymy, as distinct words sharing the same phonetic form may have different etymologies and meanings, further complicating the data selection process. One such example is the Hindi word दम /*dam*/, which can be both a word of Persian origin, meaning "spirit, life", or a Sanskrit one with a different meaning: "subdual, self-control" (McGregor 1993: 478).

The synonyms used in this analysis were selected with the following points in mind: to avoid significant differences in the definition itself; to avoid homonyms; aiming to cover a semantic field as wide and varied as possible; to include different parts of the speech; and to fulfill the definition of cognitive synonymy as defined by Cruse (2008).

## 3.2. The dataset

This study uses two corpora, namely:

- The Hindi Monolingual Corpus provided by the Indian Institute of Technology Bombay: a non-lemmatized corpus made up of roughly 768 million tokens (Kunchukuttan 2017). The origin of the texts that make up the corpus is not provided.

- As a control corpus, the hiTenTen13 corpus provided by SketchEngine, a partially lemmatized web corpus made up of texts from various newspapers and websites, comprising around 350 million words.

Synonyms analysed in both corpora were orthographically normalised. This is important for typical Perso-Arabic sounds like क़ (/q/) or ज़ (/z/), which can appear with or without nuqtā, their diacritic subscript dot (क or ज). Similarly, Unicode character combinations were standardised.

### 3.3. Word Embeddings

Word2Vec was chosen for this analysis because of its simplicity and explainability. While newer word vector tools might achieve better prediction accuracy, their results are less interpretable. In this case, we are not concerned with accuracy of prediction, but rather with analysing the contextual semantic properties of words, for which Word2Vec is an ideal fit.

Two Word2Vec models were trained for both corpora with a vector length of 200 (a widely accepted standard), a context window of five words, which means taking into account the five lexemes to the left and to the right of the target word. This is a narrow window, which seems to be appropriate to identify lexemes related by paradigmatic relations, such as synonyms; larger windows tend to be focused on semantically broader relations (Melamud et al., 2016b).

As a result, each word in the corpus is represented as a single dense vector (also known as a *word embedding*) consisting of 200 floating-point values, and the position of that word in the embedding space reflects its meaning. These vectors are generated based on the surrounding context of each word, so neighbouring words shape the word's final representation. Its semantic content is distributed across all dimensions, and each one contributes to the representation of multiple words.

Although the two corpora differ in size, both models were trained with the same value of 10 for the minimum frequency threshold. In other words, lexemes with less than 10 occurrences in the corpus were skipped (minimum frequencies of 100 and 150 were also tried, with no significant differences in classification performance).

## 3.4. Random Forest synonym classification on Word Embeddings

Classification trees function by repeatedly splitting data sets into two parts according to the most salient predictor in such a way that each split optimises classification accuracy, or any other statistical criterion—for instance, the Gini coefficient. The algorithm continues to split each resulting subset as long as statistically meaningful associations can be found between any of the predictors and the dependent variable.

Random Forest is a popular and powerful supervised classification algorithm designed as a substantial improvement on simple decision trees method. It combines a large number of independent trees trained over random and equally distributed subsets of data. Random forests select variables in individual classification trees and determine which of them are to become the most relevant predictors (Breiman 2001).

Decision Trees and Random Forests have been successfully applied to the analysis and classification of multivariate data in corpus linguistics across a wide range of fields, including language variation (Tagliamonte, Sali 2012), phonetic reduction (Dilts 2013), construction choices in spoken language (Klavan et al. 2025), syntactic variation (Szmrecsanyi et al. 2016), clause positioning (Rezaee, Golparvar, 2017), segment duration (Tomaschek et al. 2018) and the prediction of grammatical constructions (Hundt, Deshors 2018). Existing research has also addressed the potential risks and benefits of applying random forests to corpus data (Gries 2020). However, it seems that this method has not yet been used for the study of synonymy.

In this case a Random Forest model was trained on embeddings of synonym pairs in Hindi of both Sanskrit and Perso-Arabic origin, with the target value being the origin of the word.

## 3.5. Analysis of the words most prone to misclassification

To investigate mispredictions, the classifier was run cumulatively over 100,000 iterations, each time using a new random 80:20 train/test split. Cumulative misclassification counts were recorded for each word across all iterations. The most frequently misclassified words were analysed, with particular attention to their etymological origin.

Correlation matrices were then computed to examine whether other factors could influence the probability of misclassification, such as raw frequency in the corpus, signed difference in frequency between each word and its synonym, and Word2Vec cosine similarity coefficient. All experiments were repeated with the same settings on a second hiTenTen13 language corpus, which served as a control dataset. The words most frequently misclassified in both corpora were then compared.

# 4. Results

## 4.1. Random Forest Classification

The model proved to be quite successful, with an average accuracy between 79% and 95% depending on training and test data sets. The model was then trained and evaluated 100,000 times—each time on a random 80:20 train/test set—reaching an average accuracy of about 88% showing the power of the context and of the embeddings resulting from it. For each word, the cumulative sum of misclassifications was calculated, and the words were ranked according to the number of errors across all 100,000 iterations. In that cumulative run, most words were well classified, but about 20% were often prone to misclassification (Fig. 1).

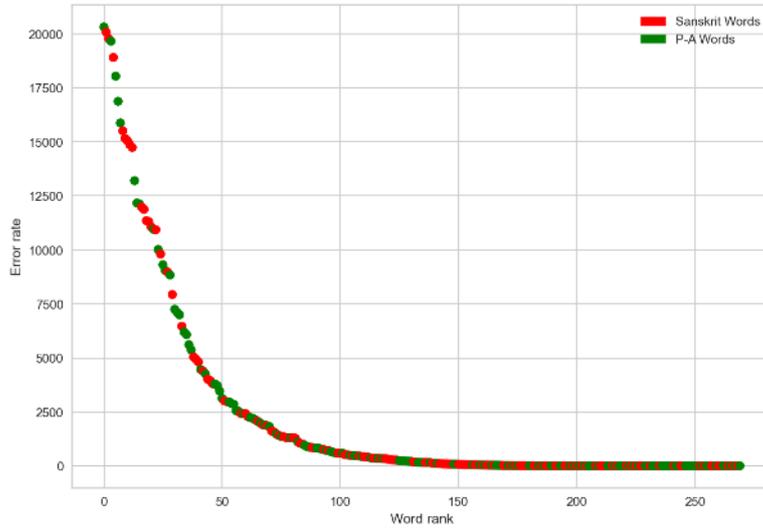

**Figure 1:** Cumulative errors after 100,000 classifier runs. Word origin is indicated by color, order is by total number of errors.

## 4.2. Tuning of the model

It has been verified what effect some key parameters of the random forest algorithm might have on the model's performance. Namely, it was checked what should be the optimal number of estimators (that is the `n_estimators` parameter in the Scikit-learn implementation) for the model, the maximum depth of each decision tree (`max_depth` parameter) and which criterion would be the best to measure the quality of split. The model was run 100 times per parameter value, and the average performance was recorded. For example, testing 200 values of n_estimators with 100 repetitions each resulted in 20,000 runs.

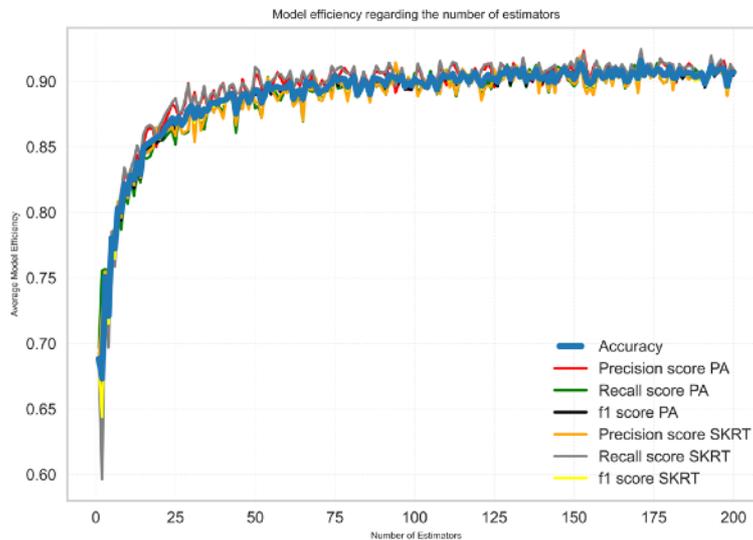

**Figure 2:** Model accuracy when trained with different number of estimators.

It seems that the number of 50 estimators is a reasonable choice, as no significant improvement in model performance was observed above that (see Fig. 2).

Similarly, the maximum depth of each decision tree was set at 5, thus preventing machine learning overfitting. Above this value, no significant changes in the model output were observed, but overfitting is thus prevented. Regarding split quality, no significant differences were noted between the Gini impurity and the Entropy measure and, finally, the Gini criterion was selected.

## 4.3. SHAP interpretation of the Random Forest results

SHAP (SHapley Additive exPlanations) coefficients are a way to understand the output of any machine learning model—including Random Forests. They allow one to measure the contribution of any single predictor on the final result by assigning it an importance value, improving the interpretability of the result.

The analysis of SHAP coefficients has shown that no predictor (i.e. no vector dimension) stands out with a particularly high impact on the model output. Even the most important predictor has limited importance in the range [-0.10, +0.06]. This indicates that the predictive power is distributed fairly evenly across the embedding space, rather than being concentrated in a few vector 'directions' (Fig. 3). However, the SHAP analysis shows clearly that some vectors in the latent space have strong bidirectional association with word origin, confirming that the embedding is capturing this information from context.

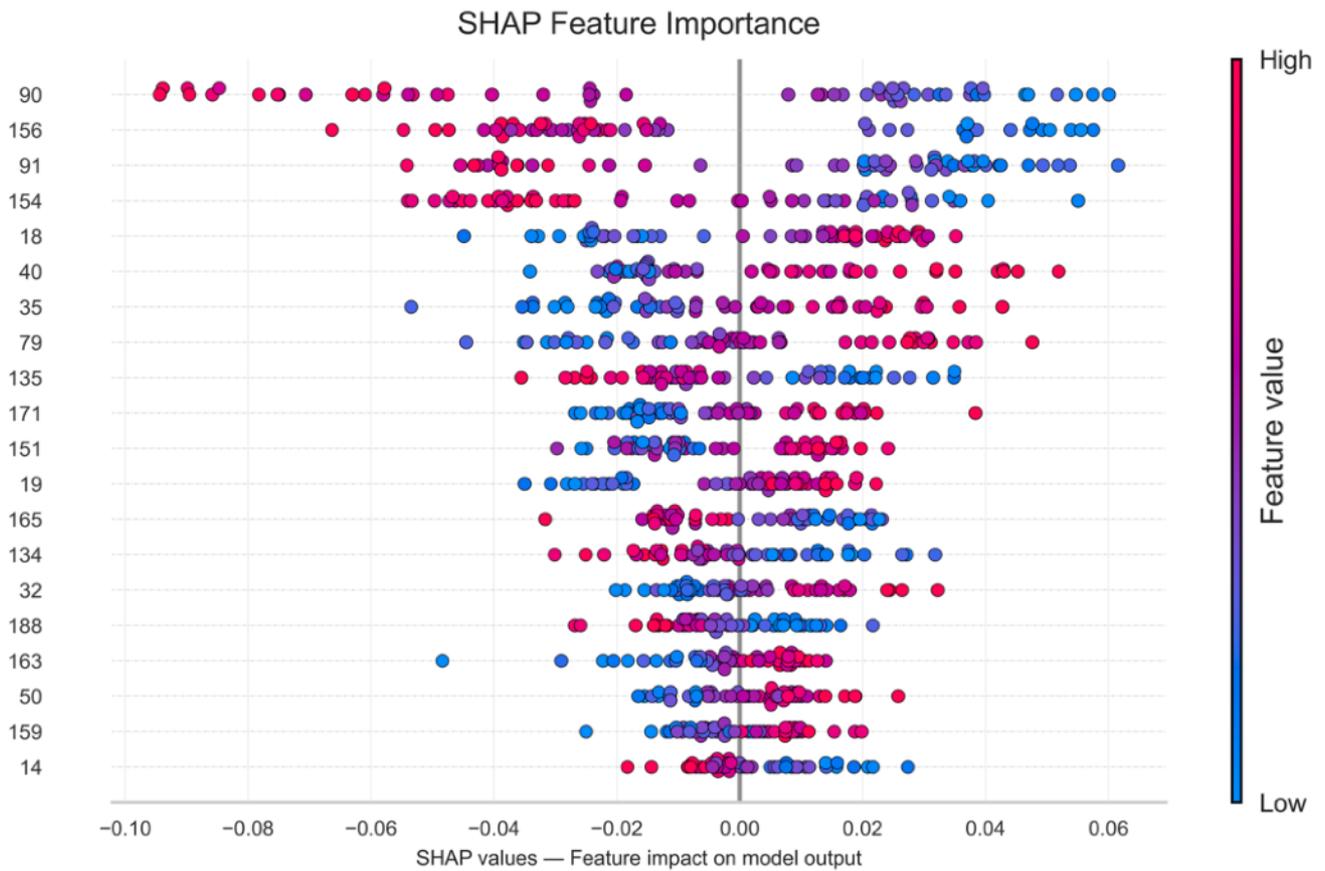

**Figure 3:** SHAP feature effects for Random Forest classification. Feature names are vector dimensions. Higher dispersion indicates greater feature effect.

## 4.4. Resilience of the model

Analysing data in high-dimensional spaces is always a challenge. Here, the number of dimensions of the word embeddings was set to 200 which is a widely accepted standard. Quite unexpectedly, it turned out that the model was able to correctly classify synonyms even when trained with limited access to all vector dimensions.

A test was conducted where different models were trained in a loop, each time using the first $n$ dimensions of the word vectors as predictors, with $n$ gradually increased from 1 to 200 (i.e., the full vector length). As before, the model was run 100 times for each value of $n$ using different train/test splits, and the average performance was recorded.

It is puzzling that, despite the low accuracy at very low values of $n$, the model started to perform well trained once it had access to at least the first 50 dimensions of the vectors, reaching an accuracy slightly above 80%. After this point the accuracy curve flattens out and grows slowly as the number of predictors, that is, the number of vector dimensions taken into account when training the model, increases (Fig. 4).

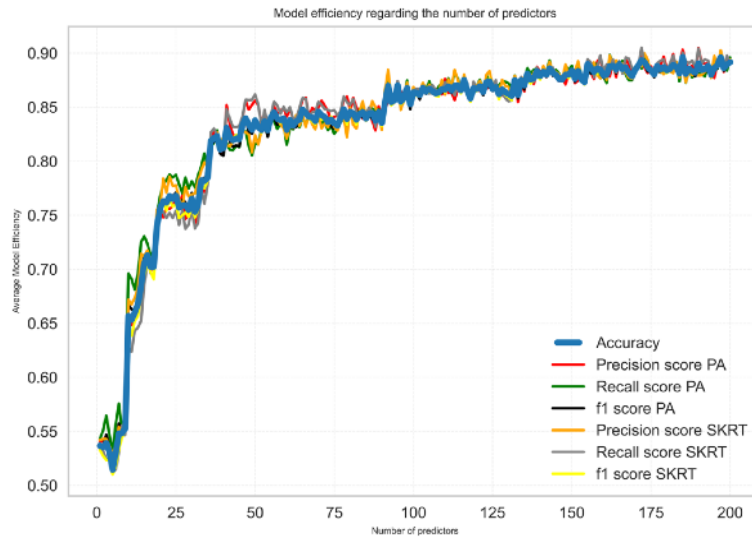

**Figure 4:** Model accuracy when trained on subsets of the full embedding space

A similar trend appeared in the second experiment, where the number of dimensions was increased in the same way, but the specific dimensions of the vectors were selected randomly (*n* random dimensions instead of first-*n*). Although in a more haphazard manner, a similar trend is also emerging here, and the model seems to stabilize from about 50 or so dimensions up (Fig. 5). This demonstrates that a lot of information of different types is condensed in the different dimensions of the vectors; this, in a sense, shows some redundancy, but also the robustness of the model despite the loss of much of the information contained in the embedding space.

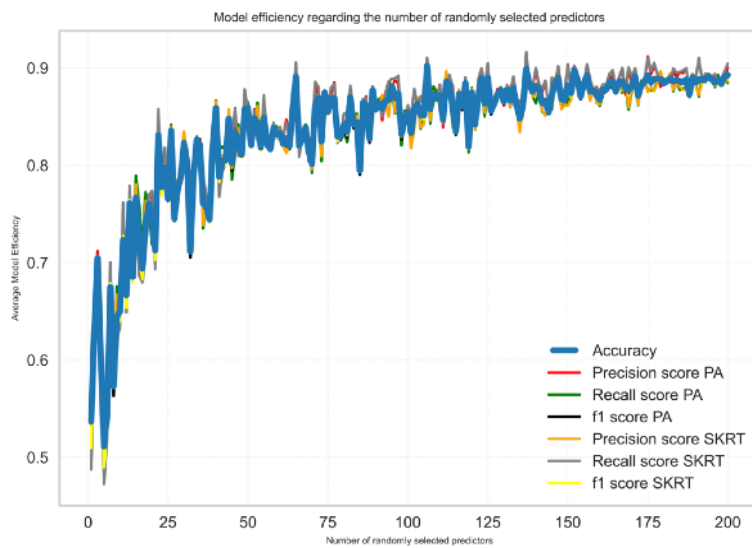

**Figure 5:** Model accuracy when trained on random subsets of the full embedding space.

Moreover, this calls for a re-examination of the types of semantic properties that are actually captured by word embeddings as they seem to go far beyond the basic lexical meaning.

## 4.5. Words most prone to misclassification

Words which were more often misclassified deserve special attention. An analysis of these words was conducted, with a focus on the origin and number of errors generated. Misclassification errors were not uniformly distributed: a relatively small set of words were responsible for most of the errors (Fig. 1).

On the one hand, many of these words are common everyday terms, such as वास्तविकता (*vāstavikata / reality*), पानी (*pānī / water*), चिट्ठी (*ciṭhṭhī / a letter*), सपना (*sapnā / dream*), as well as synsemantic words like और (*aur / and*), which are less culturally marked. In contrast, words closely associated with broad social concepts were consistently classified correctly, for example: राष्ट्र – क़ौम (*rāṣṭra – qaum / nation*), धर्म – मज़हब (*dharma – mazhab / religion*), and सेना – फ़ौज (*senā – fauj / army*). Another factor appears to be large frequency imbalances within synonymous pairs — for instance, देश – मुल्क (*desh – mulk / country*), where the former was often misclassified, while the latter was reliably recognized. With a frequency ratio of roughly 97.78% to 2.21%, this suggests that the more generic variant is harder to identify, whereas the specialized form is easier to classify. This intuition is supported by the fact that among the 20 most frequently misclassified words, the majority have a clearly positive frequency ratio — that is, they occur much more often than their counterparts.

Sanskrit words were slightly more prominent among the 20 words most prone to misclassification, comprising 58% of the total errors. However, when extended to the 50 words most prone to misclassification, the error ratio is more evenly distributed with words of Sanskrit origin responsible for about 52% of the errors. These results may vary slightly depending on the train/test datasets, but remain broadly similar.

This suggests that the origin of Perso-Arabic words is slightly more recognizable, implying that their contexts are more distinctive to the classifier, perhaps due to cultural factors. Supporting this interpretation, manual concordance analysis indicates that Sanskrit words tend to appear in broader contexts, which suggests a lower degree of specialization.

## 4.6. Random Forest Classification without words prone to misclassification

A new Random Forest model was then trained on the original dataset, excluding the 50 words that were most frequently misclassified. This model reached an average

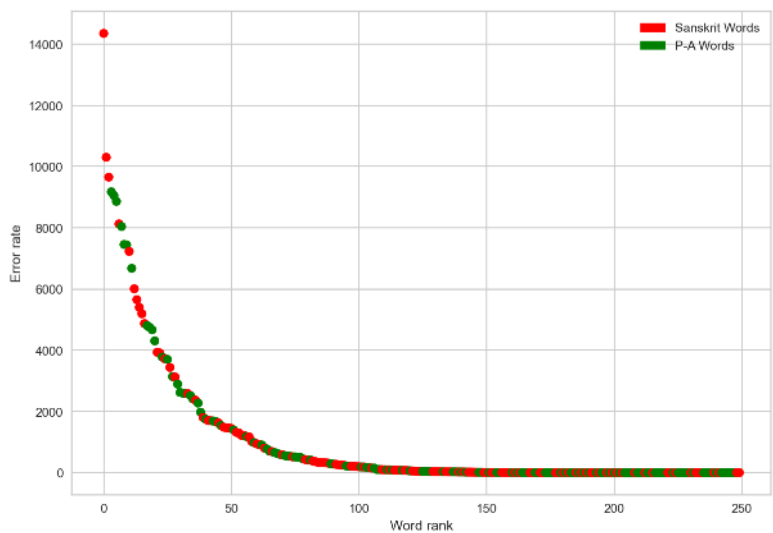

**Figure 6:** Cumulative errors after 100,000 classifier runs without the 50 most commonly misclassified words. Word origin is indicated by color, order is by total number of errors.

accuracy of 95% suggesting that the words deleted were found in unusual contexts,

while the rest were used in a more conventional fashion. As shown in Figure 6, after 100,000 iterations there are very few words still prone to misclassification.

Once again, as in the previous SHAP results, no single predictor appeared to have a particularly high impact on the model output.

## 4.7. The possible impact of word frequencies

After calculating correlation coefficients between the number of misclassifications and word frequency, it appears that less common words in the corpus are not more often misclassified, as one might expect. In fact, the opposite seems to be true: high-frequency words are more susceptible to errors, as shown in the correlation matrix (Fig. 7). Relative frequency also appears to influence misclassification: more frequent variants are more prone to errors than their less frequent counterparts. However, these correlations disappear in a model trained without the 50 words most prone to errors. By contrast, no correlation was found between misclassification likelihood and the Word2Vec cosine similarity coefficient between pairs of synonyms.

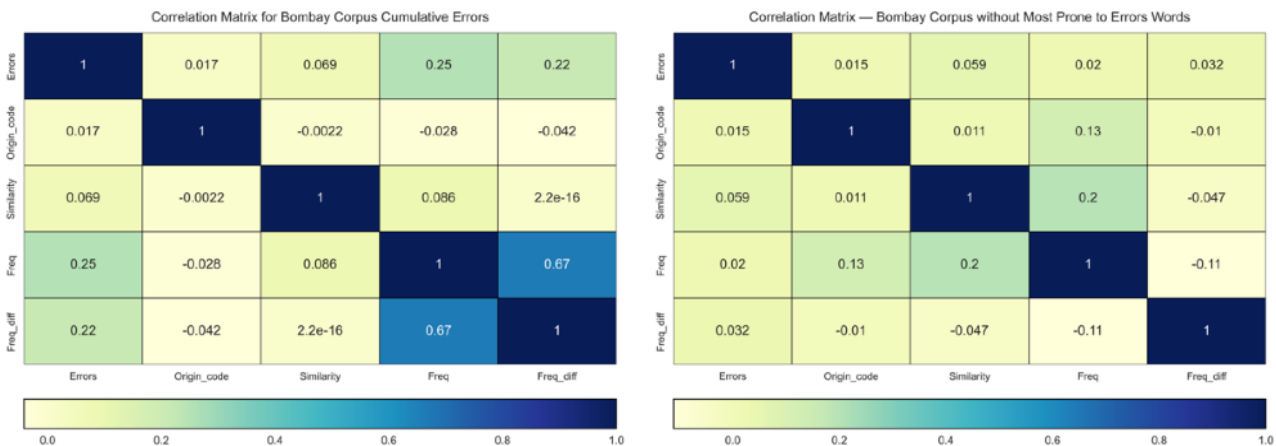

**Fig 7:** Correlation matrices with full dataset (left) and without 50 most prone to misclassification words (right).

Although frequency is not an element of language structure, it plays a major role in linguistics. Its effects have been found in virtually every subdomain of language that has been studied (Divjak 2019: 262). Thus, it would not be unusual for word frequency to also play a role in this study of synonymy.

In his pioneering research, Zipf pointed out that there is a tendency for more frequent words to be more polysemous; this is known as the Mean-frequency law. Zipf also stated that a lazy speaker would tend to use only a few highly frequent but strongly polysemous words while a more precise one would tend to use less common but highly specialized words (Zipf 1949). Combining these statements, one can assume that common, less specialized words—those with greater semantic capacity, i.e., highly polysemous—are used more frequently than highly specialized ones.

This may help explain why more frequent words seem more difficult to predict in terms of origin. Being more frequent, they appear in a wider range of contexts and are also likely more polysemous, making them less specialized. This complicates the task of predicting their origin, as their context is less characteristic. However, this observation certainly does not exhaust the entire problem of misclassification, which warrants further research.

## 4.8. Confirmation on the control corpus

To confirm the generality of the results, all experiments were repeated with the same settings on the hiTenTen13 corpus, which served as a control dataset. The words most frequently misclassified in both corpora were then compared. For the 10, 20 and 30 most misclassified words, the two sets showed considerable overlap, with only minor differences in the ranking of some words. The Jaccard index between the sets was approximately 0.43, which is remarkably high compared to the expected similarity between two randomly selected sets of words (see code repository for details).

# 5. Conclusions and Importance of the results

The key to this analysis lies precisely in the fact that the words examined were synonyms, which, prima facie, should be interchangeable. Yet the results show that even several centuries after Perso-Arabic borrowings appeared in the subcontinent, two distinct lexical strata remain discernible. A classifier based solely on word embeddings correctly categorized them by origin, indicating that the model accurately inferred the etymology of previously unseen words from vector representations of other lexemes sharing the same origin — even when they were semantically unrelated.

Since embeddings condense information about word's contextual use, it can be inferred that context—i.e., usage patterns—is sufficient to identify cultural origins that may be unknown to language users, or at least to detect sociolinguistic phenomena such as social-class distinctions, for which cultural origin often serves as a proxy. Moreover, the model was able to achieve this even without access to full-dimensional data, indicating that the effect is robust and systematic.

These findings support the idea that synonyms from different language strata exhibit distinct usage patterns, providing quantifiable evidence that a word's origin can impart cultural connotations alongside subtle distinctions between semantics and pragmatics. They suggest that etymologically related words, which may not be directly connected in a traditional semantic sense, can nonetheless form distinct networks of concepts and occupy slightly different semantic niches, sharing nuanced meanings shaped by their common origin. These results could also point toward a broader tentative cognitive or conceptual structure, which we may compare to a kind of semantic frame, where a word's meaning is understood in relation to a wide range of knowledge concerning its origin and the specific cultural background it involves.

Thus, the findings shed new light on what Divjak has aptly stated: "even if near-synonyms do name one and the same thing, they name it in different ways: they present different perspectives on a situation" (Divjak 2010: 1). This study confirms that intuition with empirical evidence and offers a foundation for further exploration —both to verify such perspectival differences and to understand how and why they arise. It also contributes to the discussion of the „No Synonymy / No Equivalence" principle, enriching it with statistical data.

The study further suggests that words from certain lexical layers — in this case, Sanskrit—may be somewhat less recognizable to the classifier. Misclassification also appears to increase slightly with word frequency and differences in occurrence counts between synonyms also seem to influence misclassification, in contrast to cosine similarity, which appears to have none. This may be attributed to the greater polysemy of high-frequency words and the fact that more specialized words from a given stratum (and thus, less frequent) occupy more distinctive semantic niches. Cultural factors may also play a role, as culturally unmarked words seem harder to classify. Therefore, both frequency and cultural factors appear to influence misclassification and thus warrant further investigation.

Finally, these results may call for a re-examination of the kinds of semantic and pragmatic properties that are actually captured by word embeddings. As this study shows, embeddings encode information that allows for predictions extending far beyond basic lexical meaning. In other words, what additional layers of meaning or social context might be encoded within the linguistic context — and therefore, within word embeddings themselves?

It would also be valuable to extend this research to other languages using the same methodology. For example, applying it to the phenomenon of synonymy in English, where Norman French and Old English lexical layers coexist, could yield particularly interesting results.

# 6. Availability of Data and Code

The analysis code may be found at https://github.com/JacekBakowski/distributive-semantics/tree/main/papers/2025-poc. All code and data is available under CC-BY, except where restricted by upstream licenses. The code repository includes full reproduction data and code for all tables and figures in this paper, as well as various supplemental figures and explanations.

# Acknowledgments

This work was funded by the National Science Centre, Poland, grant number 2021/43/O/HS2/02392.